\documentclass[a4paper,12pt]{article}  
\usepackage{graphicx}                   
\usepackage{amsmath,amsfonts,amssymb}   
\usepackage{hyperref}                   
\usepackage{geometry}                   
\usepackage{authblk}                    
\usepackage{cite}                       
\usepackage{url}                        
\usepackage{xcolor}                     
\usepackage{subcaption}
\usepackage{listings}

\title{Developing Gridded Emission Inventory from High-Resolution Satellite Object Detection for Improved Air Quality Forecasts}

\author[1]{Shubham Ghosal}
\author[2]{Manmeet Singh}
\author[3]{Sachin Ghude}
\author[2]{Harsh Kamath}
\author[3]{Vaisakh SB}
\author[3]{Subodh Wasekar}
\author[3]{Anoop Mahajan}
\author[2]{Hassan Dashtian}
\author[2]{Zong-Liang Yang}
\author[2]{Michael Young}
\author[2]{Dev Niyogi}

\affil[1]{Banaras Hindu University, Varanasi, India}
\affil[2]{The University of Texas at Austin, Austin, Texas, USA}
\affil[3]{Indian Institute of Tropical Meteorology, Pune, India}

\affil[ ]{\texttt{shubhamghosal100@gmail.com}, \texttt{\{manmeet.singh, harsh.kamath, dashtian, liang, michael.young, dev.niyogi\}@utexas.edu}, \texttt{\{sachinghude, vaisakh.sb, subodhwasekar, anoop\}@tropmet.res.in}}

\date{}

\begin{document}
\maketitle


\begin{abstract}
This study presents an innovative approach to creating a dynamic, AI-based emission inventory system for use with the Weather Research and Forecasting model coupled with Chemistry (WRF-Chem), designed to simulate vehicular and other anthropogenic emissions at satellite-detectable resolution. The methodology leverages state-of-the-art deep learning-based computer vision models, primarily employing YOLO (You Only Look Once) architectures (v8-v10) and T-Rex, for high-precision object detection. Through extensive data collection, model training, and fine-tuning, the system achieved significant improvements in detection accuracy, with F1 scores increasing from an initial 0.15 at 0.131 confidence to 0.72 at 0.414 confidence. A custom pipeline converts model outputs into netCDF files storing latitude, longitude, and vehicular count data, enabling real-time processing and visualization of emission patterns. The resulting system offers unprecedented temporal and spatial resolution in emission estimates, facilitating more accurate short-term air quality forecasts and deeper insights into urban emission dynamics. This research not only enhances WRF-Chem simulations but also bridges the gap between AI technologies and atmospheric science methodologies, potentially improving urban air quality management and environmental policy-making. Future work will focus on expanding the system’s capabilities to non-vehicular sources and further improving detection accuracy in challenging environmental conditions.
\end{abstract}

\section{Introduction}
Urban areas are experiencing rapid population growth, which has led to increased traffic congestion and vehicular emissions, becoming major contributors to air pollution. Traditionally, emission inventories—crucial for air quality modeling—are compiled using static methodologies such as periodic surveys and manual data collection. These conventional methods lack the temporal and spatial granularity necessary to capture the dynamic nature of urban emissions, especially in rapidly changing environments. As a result, emission inventories often fail to accurately represent real-time emission patterns, limiting the accuracy and responsiveness of air quality models.

A major limitation of current static emission inventories is their inability to provide daily or sub-daily data as they are developed for particular years and are reused after that. This temporal lag results in air quality prediction models that cannot capture short-term fluctuations in vehicular traffic or sudden changes in emission levels. This restricts the ability of air quality models to provide realistic representations of urban pollution, thereby reducing their effectiveness for forecasting and policy-making.

\begin{figure}[h]
    \centering
    \fbox{\includegraphics[width=0.6\textwidth]{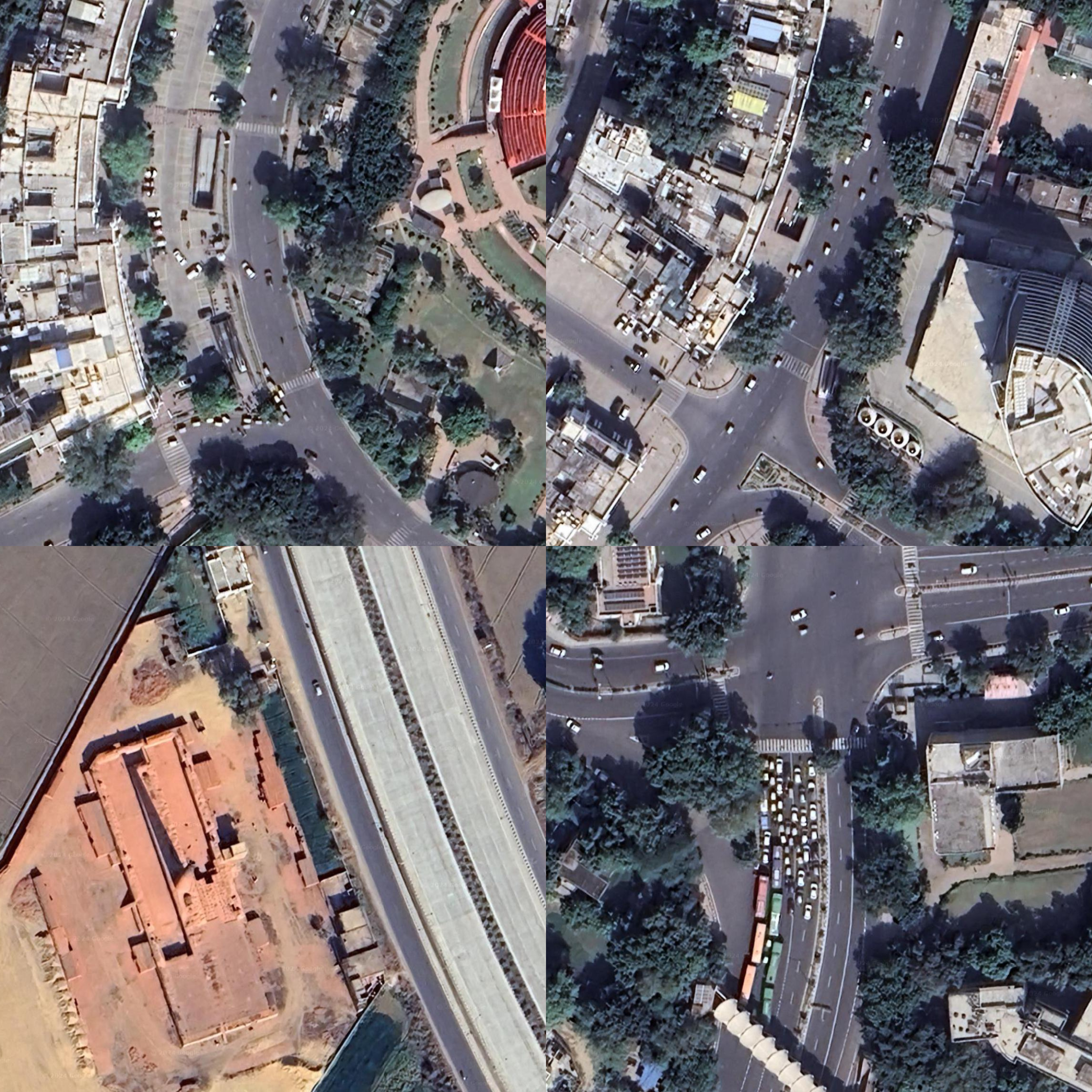}}
    \caption{ Satellite image of Cannaught place, Delhi with cars, buses, trucks, brick kilns and other visible sources.
CNES/Airbus, Maxar Technologies, Copyright Google 2024}
    \label{fig:image_label}
\end{figure}

This study addresses the limitations of static emission inventories by developing an AI-driven system capable of generating real-time, high-resolution emission data using satellite imagery. By leveraging state-of-the-art deep learning models such as YOLOv10 for object detection, this approach enables continuous monitoring of vehicle counts and classification, linking them directly to emission factors. The resulting system allows for dynamic emission inventories that can feed into air quality models on daily or sub-daily timescales, thereby improving the accuracy and timeliness of air quality forecasts.The objective of this study is to develop an AI-based emission inventory system using cutting-edge deep learning object detection models to detect vehicles, classify them by type, and link them to emission factors to derive real-time estimations of pollution levels. This paper presents a novel methodology for tracking vehicular emissions using satellite images, aiming to provide accurate, timely, and spatially-resolved data for urban traffic management and environmental monitoring. We used geotagged .tiff files processed through the YOLOv10 object detection framework, along with other viable methods, to identify vehicles such as cars, buses, and brick kilns. These detected objects were then linked to corresponding emission factors to produce vehicle counts over the region, providing a more dynamic and accurate emission inventory.

\subsection{Aim \& Contribution}
This study represents a significant advancement in the field of emission inventory generation by utilizing very high-resolution satellite imagery to develop the first-ever gridded vehicle count system. The primary aim of this work is to create a dynamic, AI-driven method that accurately tracks and counts vehicles in urban environments, laying the foundation for future developments in real-time emission inventories. By combining cutting-edge deep learning techniques with satellite data, this system addresses the limitations of static emission inventories and provides a more temporally and spatially resolved dataset. The key contributions of this study are outlined below.

\subsection{First-Time Development of Gridded Vehicle Count from High-Resolution Satellite Imagery}
This work pioneers the creation of a gridded vehicle count system using very high-resolution satellite imagery. This gridded dataset provides a comprehensive view of vehicular activity in urban areas, with each grid cell containing detailed vehicle count information. This breakthrough allows for the development of more dynamic, real-time emission inventories that can be updated frequently, unlike traditional methods that rely on periodic, static data.

\subsection{Accurate Foundation for Future Emission Inventory Generation}
The gridded vehicle count system offers a crucial foundation for creating highly accurate emission inventories. By linking the detected vehicles in each grid cell to their respective emission factors, future emission models can be built with far greater precision. This approach enables the continuous monitoring of urban vehicular emissions, paving the way for more realistic and responsive air quality models that reflect daily or sub-daily changes in traffic patterns.

\subsection{Multiple Classes in Gridded Vehicle Count}
The system categorizes the detected vehicles into four main classes: brick kilns, cars, buses, and other vehicles (such as motorcycles, auto-rickshaws, and smaller vehicles). This classification system allows for a more nuanced understanding of emission sources, as each class can be linked to specific emission factors. The ability to distinguish between different vehicle types enhances the accuracy of the resulting emission inventories and helps identify high-emission sources more precisely.

\subsection{Development of Labeled Data from Scratch}
A significant contribution of this work is the development of a labeled dataset, created entirely from scratch. The high-resolution satellite imagery used in this study was manually annotated to identify vehicles and other relevant objects. This effort involved labeling tens of thousands of images, which were then used to train deep learning models like YOLOv10 for object detection. The resulting labeled dataset is a valuable resource for future research and model training, enabling the continued refinement of vehicle detection and classification methods.

Overall, this study lays the groundwork for a more advanced approach to emission inventory generation, leveraging AI and satellite imagery to provide real-time, high-resolution data that can significantly improve urban air quality models and policy-making efforts.

\section{Methodology}

The methodology consists of several key steps, each of which is critical to achieving accurate results for real-time emission inventory generation:

\begin{figure}[h]
	\centering
	\fbox{\includegraphics[width=0.65\textwidth]{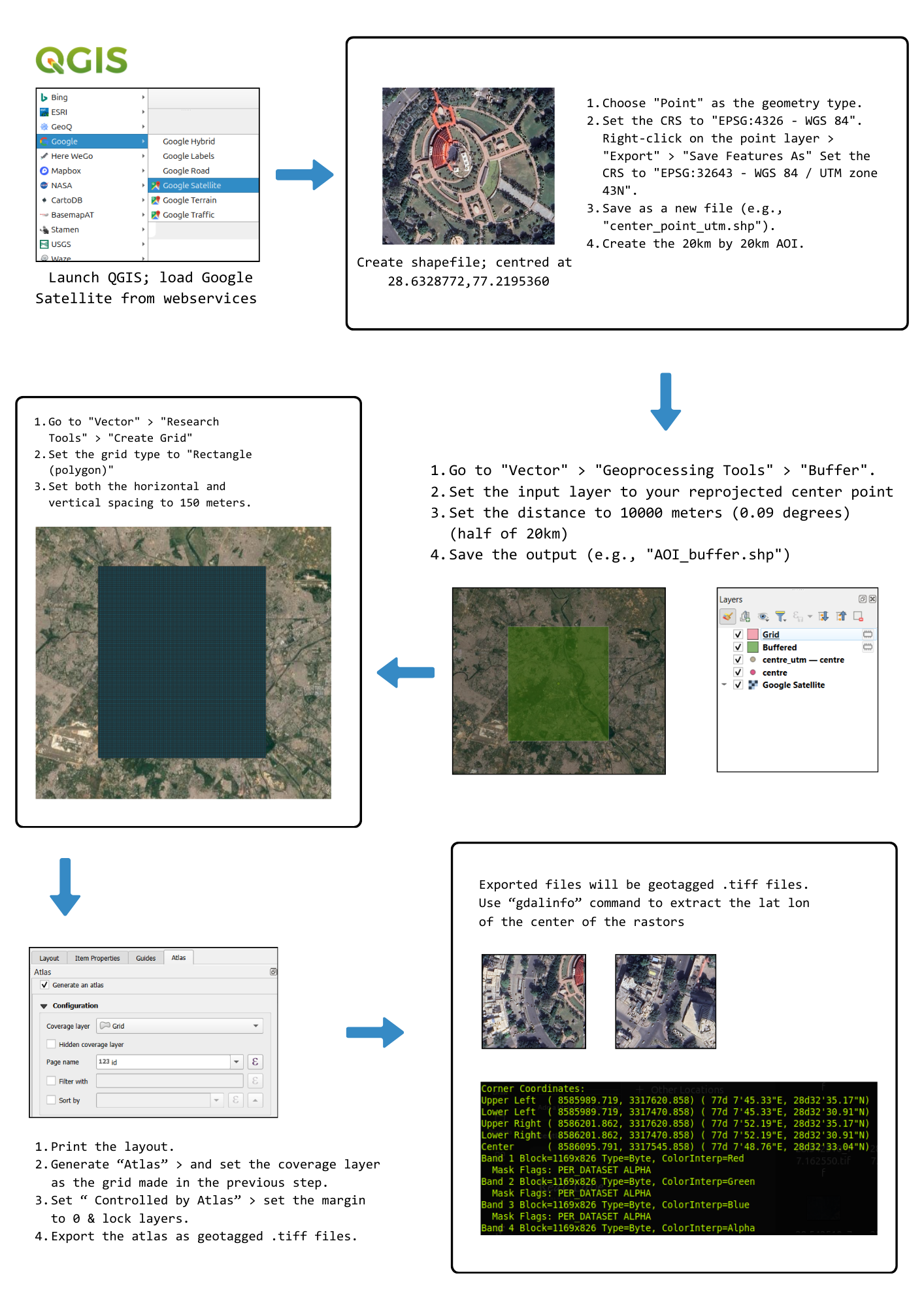}}
	\caption{Methodology Flowchart for Data Collection}
\end{figure}

\subsection{Data Collection}
High-resolution satellite imagery was sourced for an extensive study focused on urban environments, specifically targeting the Connaught Place region in New Delhi. To facilitate systematic analysis, the geographic region of interest (ROI) was defined, and the data collection process was meticulously organized using QGIS software. The first step involved loading Google Satellite layers into QGIS to capture detailed, high-resolution satellite images of specific urban locations. Using QGIS's \textit{Atlas} feature, we generated geotagged \texttt{.tiff} files for each selected area. This ensured that the files not only captured the visual information from the satellite but were also embedded with geographic metadata, including latitude and longitude coordinates.

\begin{figure}[h]
	\centering
	\begin{subfigure}[b]{0.4\textwidth}
		\centering
		\fbox{\includegraphics[height=0.25\textheight]{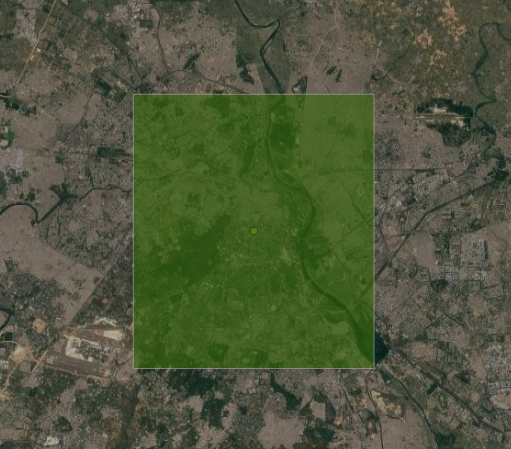}}
		\caption{Area of Interest}
	\end{subfigure}
	\hspace{1cm}
	\begin{subfigure}[b]{0.4\textwidth}
		\centering
		\fbox{\includegraphics[height=0.25\textheight]{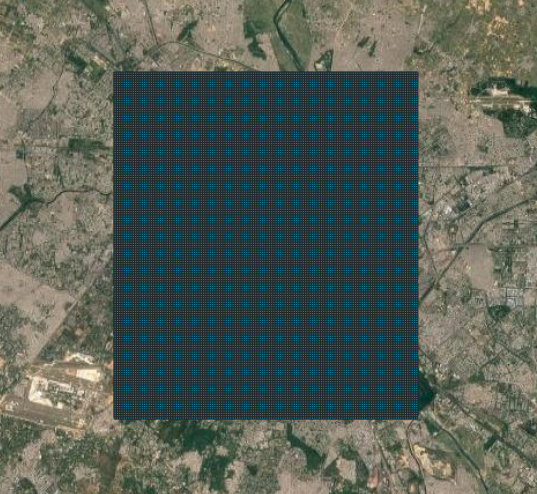}}
		\caption{Grids}
	\end{subfigure}
	\caption{Data Collection - Satellite Images}
\end{figure}

Before exporting these geotagged images, essential preparatory steps were taken to ensure the data could be analyzed efficiently. A series of shapefiles were created to establish a grid system for the study area. The grid spacing was configured with both horizontal and vertical intervals set at 150 meters, providing a structured framework for evaluating vehicle density across the region. This grid system covered a 20~km\textsuperscript{2} area, enabling an accurate and comprehensive analysis of the urban landscape. Additionally, QGIS’s geoprocessing tools were utilized to precisely center the ROI and apply a buffer zone around the study area, ensuring a consistent focus on the desired 20~km\textsuperscript{2} section of New Delhi. The shapefiles were integral to delineating this area and managing the data processing flow.

In total, 20,503 individual \texttt{.tiff} files were generated, each containing not only the high-resolution imagery but also essential geotagging metadata, such as the central latitude and longitude coordinates of each grid cell. Each file also held detailed format information, including band details that could be useful in various post-processing tasks, such as analyzing spectral data or refining the object detection algorithms. This comprehensive approach allowed for precise, large-scale image collection, paving the way for subsequent vehicle detection and density analysis.

\subsection{Data Annotation and Model Training}
The images were annotated using \textcolor{blue}{\href{https://app.roboflow.com/}{Roboflow}}, an online platform for image annotation and dataset management. In this process, vehicles were manually identified, and bounding boxes were carefully drawn around various objects such as cars, buses, and other identified entities. The workflow followed for this project was straightforward: it began with creating a new project in Roboflow, followed by uploading a subset of the collected \texttt{.tiff} files.

Once the imagery was uploaded, annotations were initiated. For this study, four distinct object classes were defined: \textit{cars}, \textit{buses}, \textit{miscellaneous} (which includes two-wheelers, auto-rickshaws, and similar small vehicles), and \textit{brick kilns}. Each of these classes was annotated by drawing bounding boxes around the corresponding objects within the satellite images. Upon completing the annotation process, the annotated dataset was automatically split into three subsets: training, validation, and testing sets. This division was essential to ensure robust model training, validation of the learned parameters, and performance testing in unseen conditions.

The dataset was then prepared for training a YOLOv10 model tailored for vehicle detection. Custom dataset configurations were applied to account for region-specific vehicle types and detection conditions that are characteristic of the urban environment under study. This ensured that the detection model would be effective in identifying the wide range of vehicle types and infrastructure components typical in the region. Upon finalizing the dataset, \texttt{.yaml} configuration files were generated. These files, which contain detailed metadata about the dataset and its class structure, were used during the training process to ensure proper class mappings and dataset handling.

\begin{verbatim}
train: ../train/images
val: ../valid/images
test: ../test/images

nc: 4
names: ['brick_kilns', 'bus', 'car', 'miscellaneous']
\end{verbatim}

\begin{figure}[h]
	\centering
	\begin{subfigure}[b]{0.4\textwidth}
		\centering
		\fbox{\includegraphics[height=0.25\textheight]{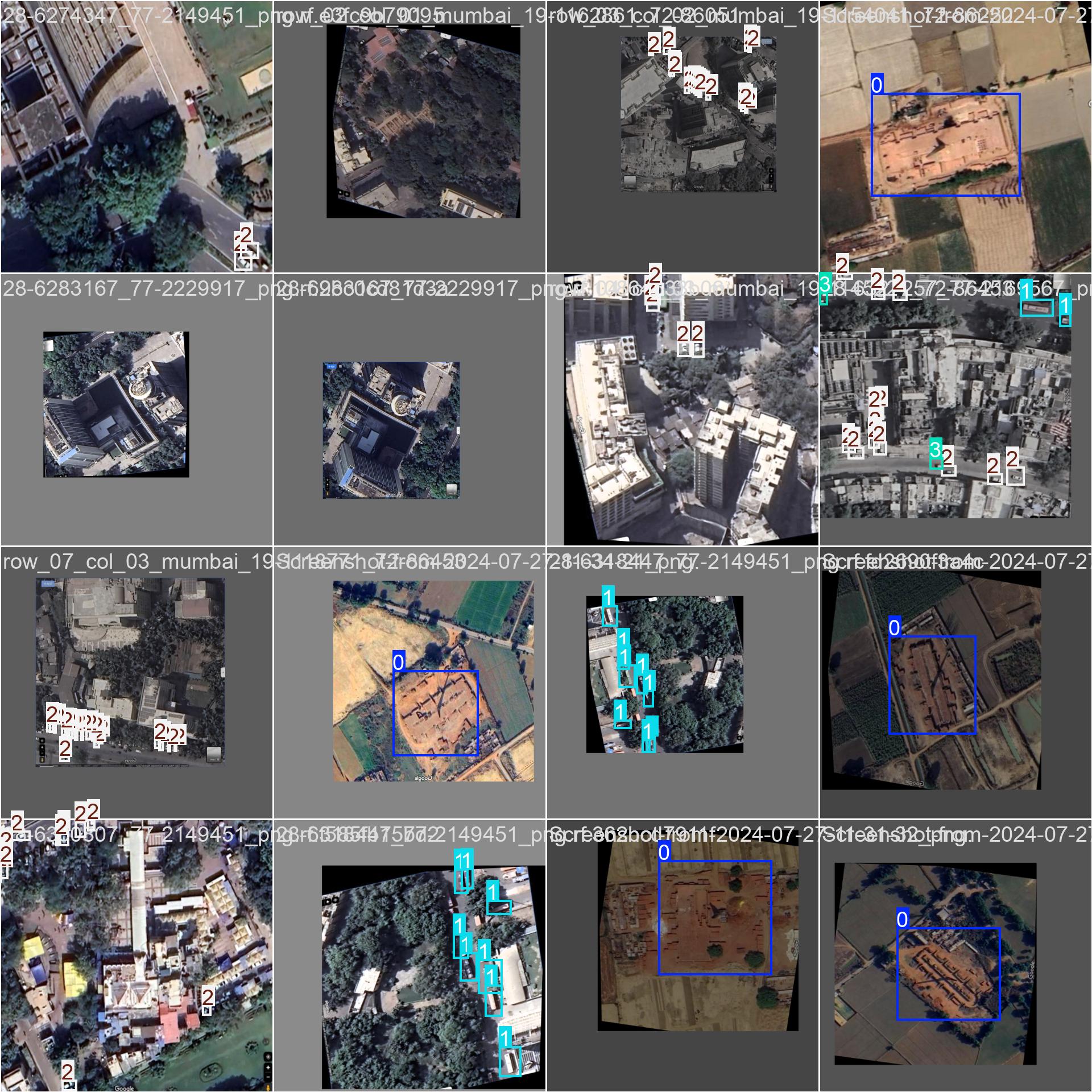}}
		\caption{Augmentation}
	\end{subfigure}
	\hspace{1cm}
	\begin{subfigure}[b]{0.4\textwidth}
		\centering
		\fbox{\includegraphics[height=0.25\textheight]{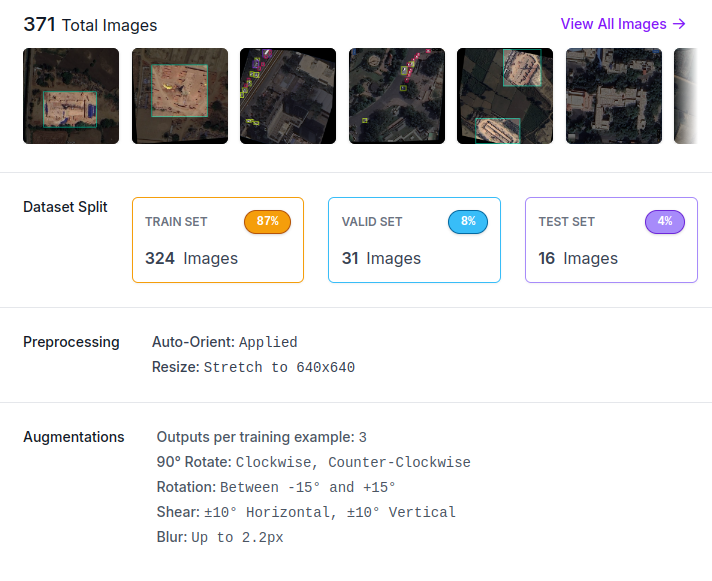}}
		\caption{Roboflow Dataset}
	\end{subfigure}
	\caption{Annotating Images for Defined Object Classes}
\end{figure}

\subsection{Vehicle Detection and Calculations}
There were three vehicle detection methods employed primarily for object detection in this study:

\begin{enumerate}
    \item \textbf{T-Rex: Counting by Visual Prompting}: T-Rex is a visual prompting-based detection method. While this approach is interactive, it relies heavily on third-party tools and is less robust for large datasets.
    
    \item \textbf{YOLOv8 to YOLOv10}: This method proved to be the most effective for the specific dataset, offering faster training and inference times compared to other models. The architectural details and tradeoffs of the YOLO versions will be discussed in later sections.
    
    \item \textbf{ChatGPT-4 Image Recognition}: Although this method is faster, it involves third-party processing, and its efficiency when handling large datasets is still limited by computational expense.
\end{enumerate}

Despite the advantages of the ChatGPT-4 method in terms of speed, YOLOv8 to YOLOv10 demonstrated the best performance overall, particularly when dealing with the large and complex datasets used in this study. Once suitable directories were created, training the model for the custom dataset was as simple as running the following code:

\begin{verbatim}
from ultralytics import YOLO
import numpy as np

# Load a model
model = YOLO("yolov10n.pt")

# Use the model
# train the model
results = model.train(data=os.path.join(ROOT_DIR, "data.yaml"), epochs=150)  
\end{verbatim}

Upon training, which is significantly faster in YOLOv10 (taking around 30 minutes for 150 epochs), running detection on any particular \texttt{.tiff} file can be performed using the following command:

\begin{verbatim}
!yolo task=detect mode=predict model='./best.pt' 
conf=.25 source='/./test_image.tiff/'
\end{verbatim}

Once detection is complete, the results for each \texttt{.tiff} file, including latitude, longitude, and object counts for each grid point, are stored in a \texttt{netCDF} file for further analysis.

\begin{figure}[h]
	\centering
	\fbox{\includegraphics[width=0.63\textwidth]{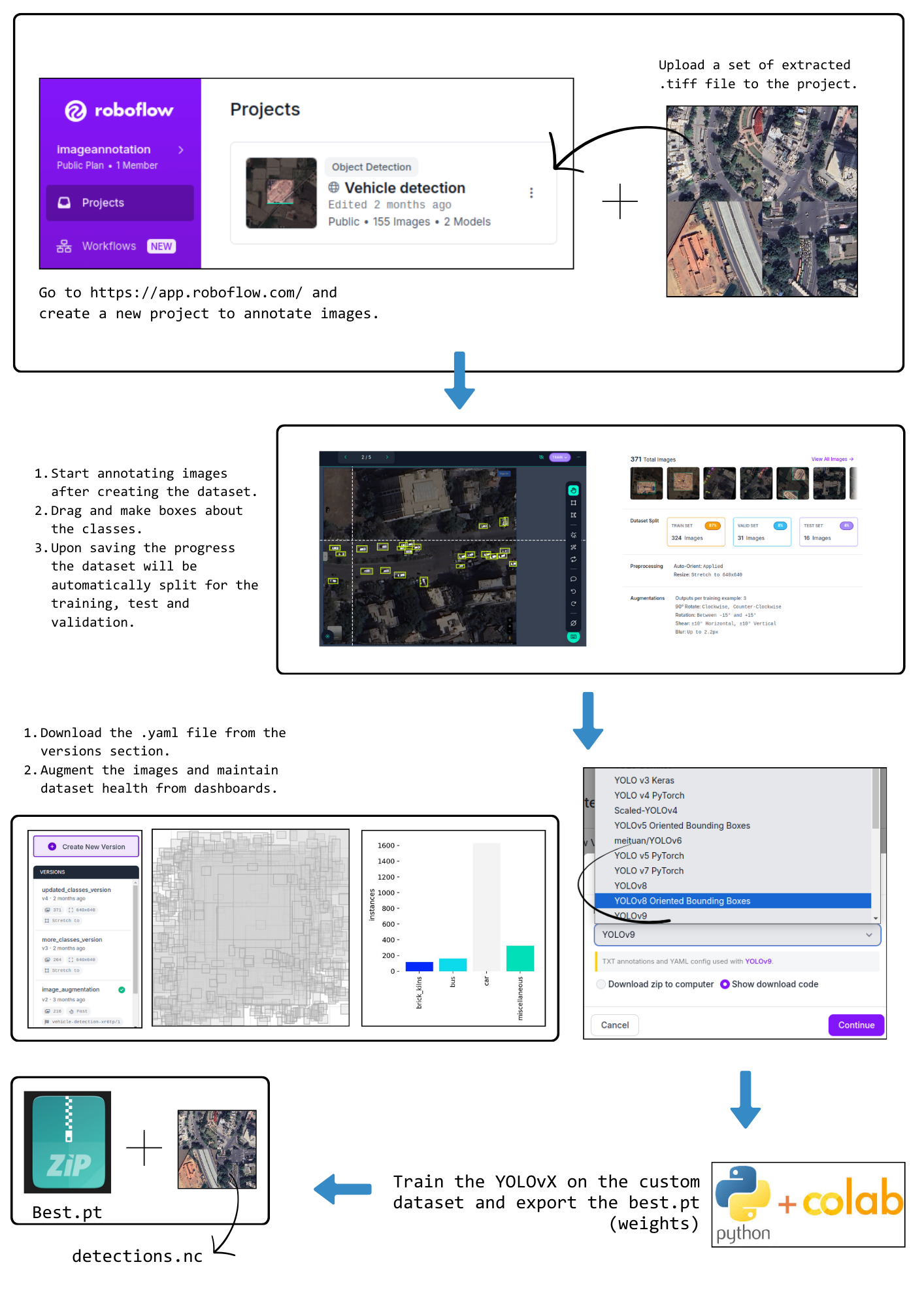}}
	\caption{Flowchart - Annotations and Model Training}
\end{figure}

\subsection{Heatmap Generation}
Using the detection results, a spatial heatmap of vehicle density was created with Matplotlib, highlighting regions with high and low traffic intensity. The final heatmap captured the cumulative vehicle detections over time, providing an accurate representation of traffic patterns. High vehicle density was observed around busy roads and intersections, while areas with vegetation or buildings were accurately reflected with lower or zero detection counts, offering a comprehensive view of the urban landscape's traffic distribution.

\section{Results}
\subsection{F1 Scores and Model Performance}

\begin{figure}[h]
	\centering
	\begin{subfigure}[b]{0.4\textwidth}
		\centering
		\fbox{\includegraphics[height=0.23\textheight]{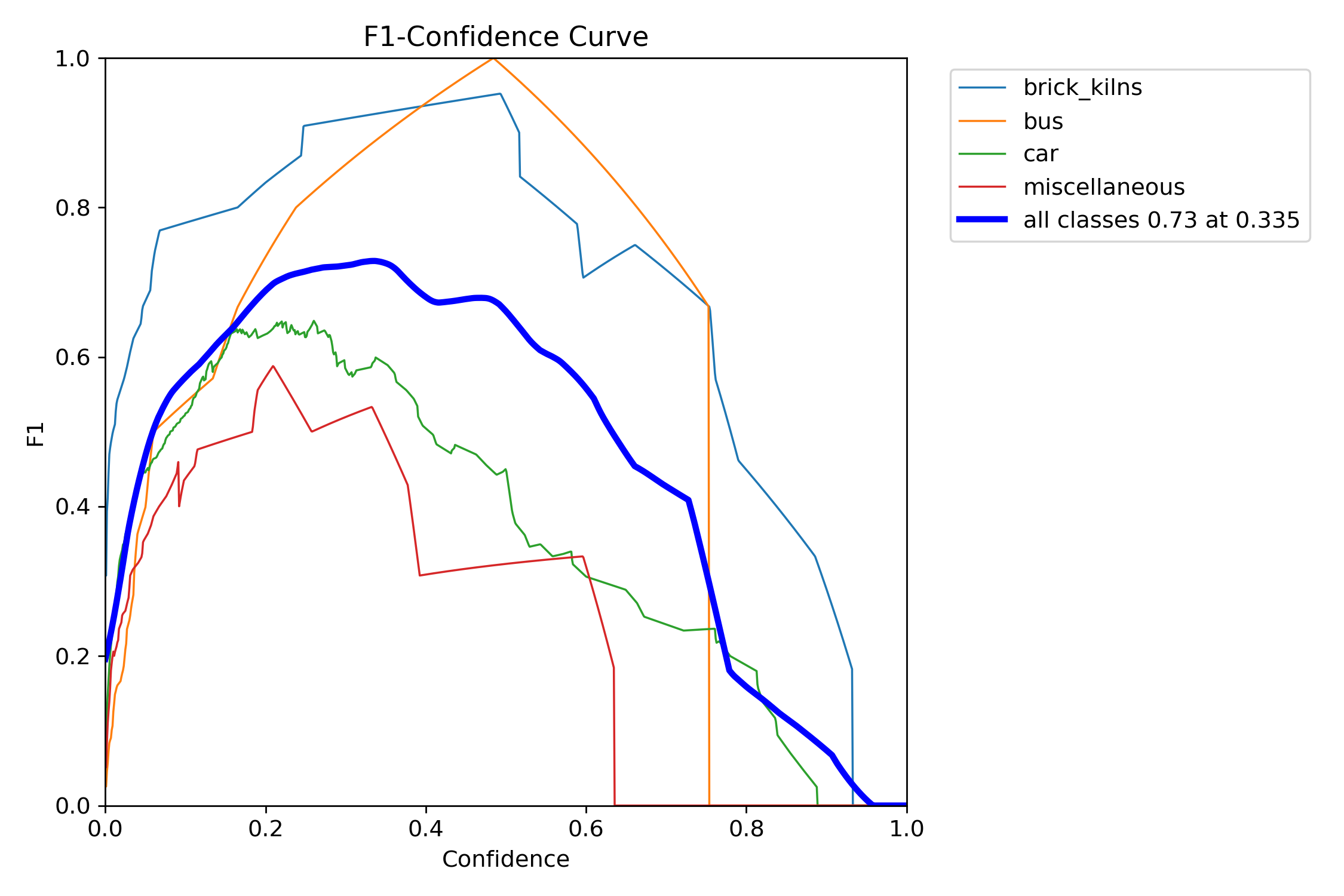}}
		\caption{F1 Confidence Curve}
		\label{fig: f1confidence}
	\end{subfigure}
	\hspace{1.8cm}
	\begin{subfigure}[b]{0.4\textwidth}
		\centering
		\fbox{\includegraphics[height=0.23\textheight]{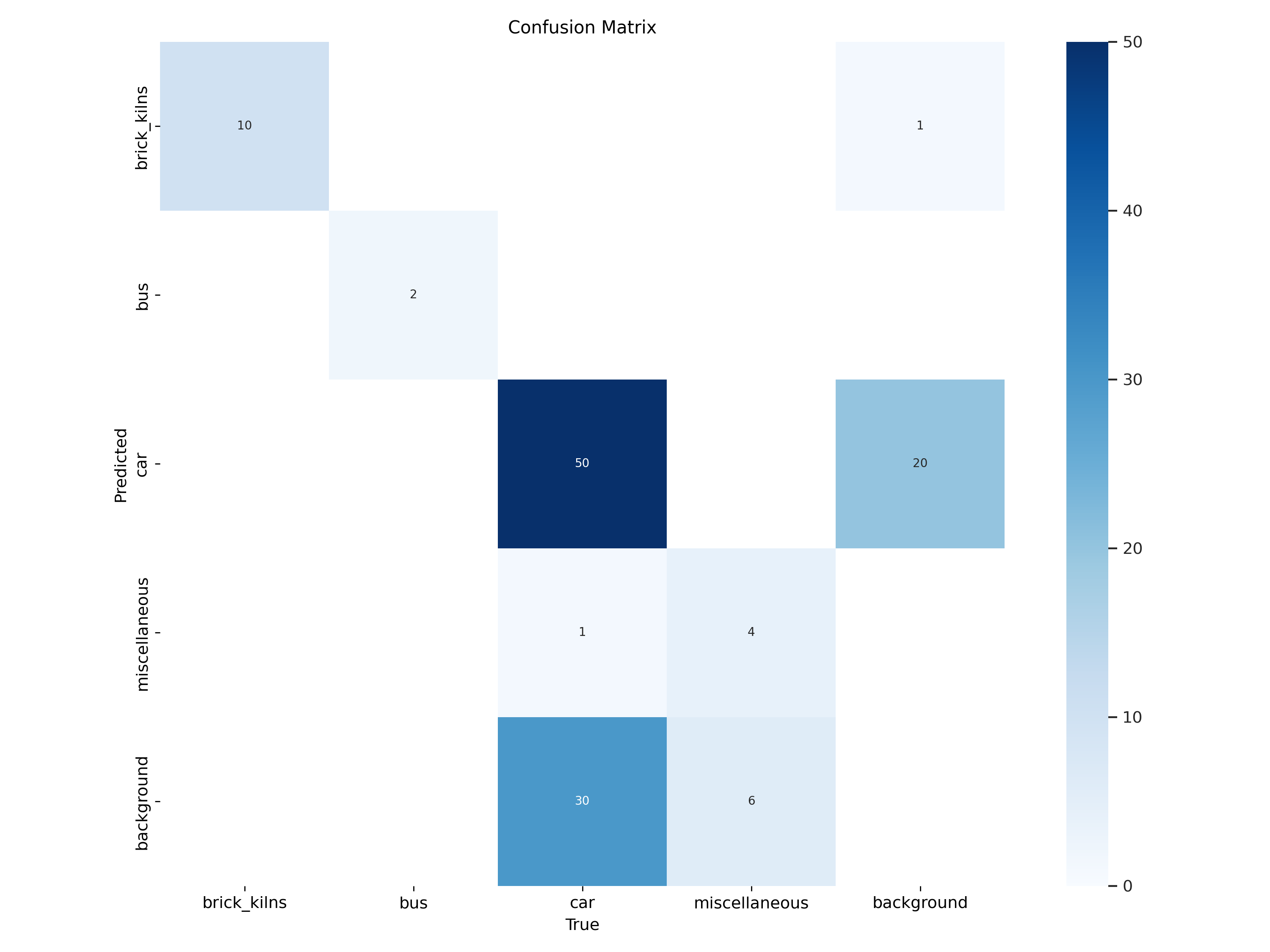}}
		\caption{Confusion Matrix}
		\label{fig: confmatrix}
	\end{subfigure}
	\caption{YOLOv10 Metrics}
\end{figure}

The trained model's performance was evaluated using confusion matrices and F1-confidence curves, as shown in Figures \textcolor{blue}{\ref{fig: f1confidence}} and \textcolor{blue}{\ref{fig: confmatrix}}. These figures provide insights into the detection accuracy and robustness of the YOLOv10 model across the four classes.

As evident from the F1-Confidence Curve in Figure \textcolor{blue}{\ref{fig: f1confidence}}, the overall F1 score for all classes peaked at 0.73 when the confidence threshold was set to 0.335, indicating high accuracy for vehicle detection, particularly for cars. While other classes, such as buses and miscellaneous objects, achieved lower true positive rates, the YOLOv10 model performed well for the majority of the dataset.

\subsection{YOLOv8 and YOLOv10: Key Comparisons}

To provide a clearer understanding of the advancements in YOLOv10, a comparison with YOLOv8 is essential. Below are some of the key improvements:

\begin{table}[h!]
\centering
\begin{tabular}{|p{4cm}|p{6cm}|p{6cm}|}
\hline
\textbf{Aspect} & \textbf{YOLOv8} & \textbf{YOLOv10} \\
\hline
\textbf{Architectural Efficiency} & 
- Utilizes C2f building block for feature extraction and fusion. \newline
- Relies on NMS for post-processing. & 
- NMS-free architecture with dual assignments. \newline
- Lightweight classification head reduces computational redundancy. \\
\hline
\textbf{Inference Speed \& Latency} & 
- Fast inference speed. \newline
- Slightly hindered by reliance on NMS, increasing latency. & 
- Faster post-processing due to NMS-free design. \newline
- YOLOv10-S is 1.8× faster than RT-DETR-R18 under similar conditions. \\
\hline
\textbf{Detection Performance} & 
- Good performance across object detection tasks. \newline
- Struggles with small objects, often requiring confidence tuning. & 
- Superior at detecting small objects, especially at lower confidence thresholds. \newline
- Dual assignment strategy ensures more consistent detections. \\
\hline
\textbf{Parameter Optimization} & 
- Efficient but has room for improvement in parameter utilization. & 
- Optimized usage, achieving higher performance with fewer parameters. \newline
- 46\% lower latency and 25\% fewer parameters compared to YOLOv9-C for similar performance. \\
\hline
\end{tabular}
\caption{Comparison of YOLOv8 and YOLOv10 across various aspects.}
\end{table}

\subsection{Key Takeaways}

\begin{itemize}
  
    \item \textbf{Speed and Efficiency}: YOLOv10 outperforms YOLOv8 in terms of post-processing speed, making it ideal for real-time applications where low latency is critical.
    \item \textbf{Detection Accuracy}: Both models perform well, but YOLOv10 excels in detecting smaller objects and offers more flexibility with confidence thresholds.
    \item \textbf{Parameter Utilization}: YOLOv10 is more compact and efficient, making it a suitable choice for applications that require both high accuracy and fast inference.
\end{itemize}

\section{Discussion}
This study highlights the potential of integrating AI with satellite imagery for real-time emission monitoring in urban environments. The approach is highly scalable and can be adapted to other urban areas, providing valuable insights into traffic patterns, emissions, and air quality. These results are particularly beneficial for urban planners, policymakers, and environmental agencies, offering a detailed tool for monitoring and mitigating vehicular emissions at a granular level. The description of the collected \texttt{.tiff} data over which the deep learning model has been trained and run on to create the gridded vehicular counts can be detailed using \texttt{gdalinfo} command:

\begin{verbatim}
>>>gdalinfo 28.542510_77.130210.tiff

Driver: GTiff/GeoTIFF
Files: 28.542510_77.130210.tiff
Size is 1169, 826
Coordinate System is:
PROJCRS["WGS 84 / Pseudo-Mercator",
    BASEGEOGCRS["WGS 84",
        ENSEMBLE["World Geodetic System 1984 ensemble",
            MEMBER["World Geodetic System 1984 (Transit)"],
            MEMBER["World Geodetic System 1984 (G730)"],
            MEMBER["World Geodetic System 1984 (G873)"],
            MEMBER["World Geodetic System 1984 (G1150)"],
            MEMBER["World Geodetic System 1984 (G1674)"],
            MEMBER["World Geodetic System 1984 (G1762)"],
            MEMBER["World Geodetic System 1984 (G2139)"],
            ELLIPSOID["WGS 84",6378137,298.257223563,
                LENGTHUNIT["metre",1]],
            ENSEMBLEACCURACY[2.0]],
        PRIMEM["Greenwich",0,
            ANGLEUNIT["degree",0.0174532925199433]],
        ID["EPSG",4326]],
    CONVERSION["Popular Visualisation Pseudo-Mercator",
        METHOD["Popular Visualisation Pseudo Mercator",
            ID["EPSG",1024]],
        PARAMETER["Latitude of natural origin",0,
            ANGLEUNIT["degree",0.0174532925199433],
            ID["EPSG",8801]],
        PARAMETER["Longitude of natural origin",0,
            ANGLEUNIT["degree",0.0174532925199433],
            ID["EPSG",8802]],
        PARAMETER["False easting",0,
            LENGTHUNIT["metre",1],
            ID["EPSG",8806]],
        PARAMETER["False northing",0,
            LENGTHUNIT["metre",1],
            ID["EPSG",8807]]],
    CS[Cartesian,2],
        AXIS["easting (X)",east,
            ORDER[1],
            LENGTHUNIT["metre",1]],
        AXIS["northing (Y)",north,
            ORDER[2],
            LENGTHUNIT["metre",1]],
    USAGE[
        SCOPE["Web mapping and visualisation."],
        AREA["World between 85.06°S and 85.06°N."],
        BBOX[-85.06,-180,85.06,180]],
    ID["EPSG",3857]]
Data axis to CRS axis mapping: 1,2
Origin = (8585989.719322871416807,3317620.858127291314304)
Pixel Size = (0.181473787118728,-0.181598062952868)
Metadata:
  AREA_OR_POINT=Area
  TIFFTAG_RESOLUTIONUNIT=2 (pixels/inch)
  TIFFTAG_XRESOLUTION=100
  TIFFTAG_YRESOLUTION=100
Image Structure Metadata:
  COMPRESSION=LZW
  INTERLEAVE=PIXEL
Corner Coordinates:
Upper Left  ( 8585989.719, 3317620.858) ( 77d 7'45.33"E, 28d32'35.17"N)
Lower Left  ( 8585989.719, 3317470.858) ( 77d 7'45.33"E, 28d32'30.91"N)
Upper Right ( 8586201.862, 3317620.858) ( 77d 7'52.19"E, 28d32'35.17"N)
Lower Right ( 8586201.862, 3317470.858) ( 77d 7'52.19"E, 28d32'30.91"N)
Center      ( 8586095.791, 3317545.858) ( 77d 7'48.76"E, 28d32'33.04"N)
Band 1 Block=1169x826 Type=Byte, ColorInterp=Red
  Mask Flags: PER_DATASET ALPHA 
Band 2 Block=1169x826 Type=Byte, ColorInterp=Green
  Mask Flags: PER_DATASET ALPHA 
Band 3 Block=1169x826 Type=Byte, ColorInterp=Blue
  Mask Flags: PER_DATASET ALPHA 
Band 4 Block=1169x826 Type=Byte, ColorInterp=Alpha
\end{verbatim}

The adaptability of this method lies in its ability to detect various object classes (e.g., industrial sources, brick kilns) and its flexibility to extend across different regions. It can also integrate more advanced machine learning models to enhance accuracy. Future work could involve coupling this inventory with air quality models to predict pollution levels based on real-time traffic data. Data collection for this study was conducted using a QGIS plugin connected to Google satellite imagery. 

\begin{figure}[h]
	\centering
	\fbox{\includegraphics[height=0.68\textheight]{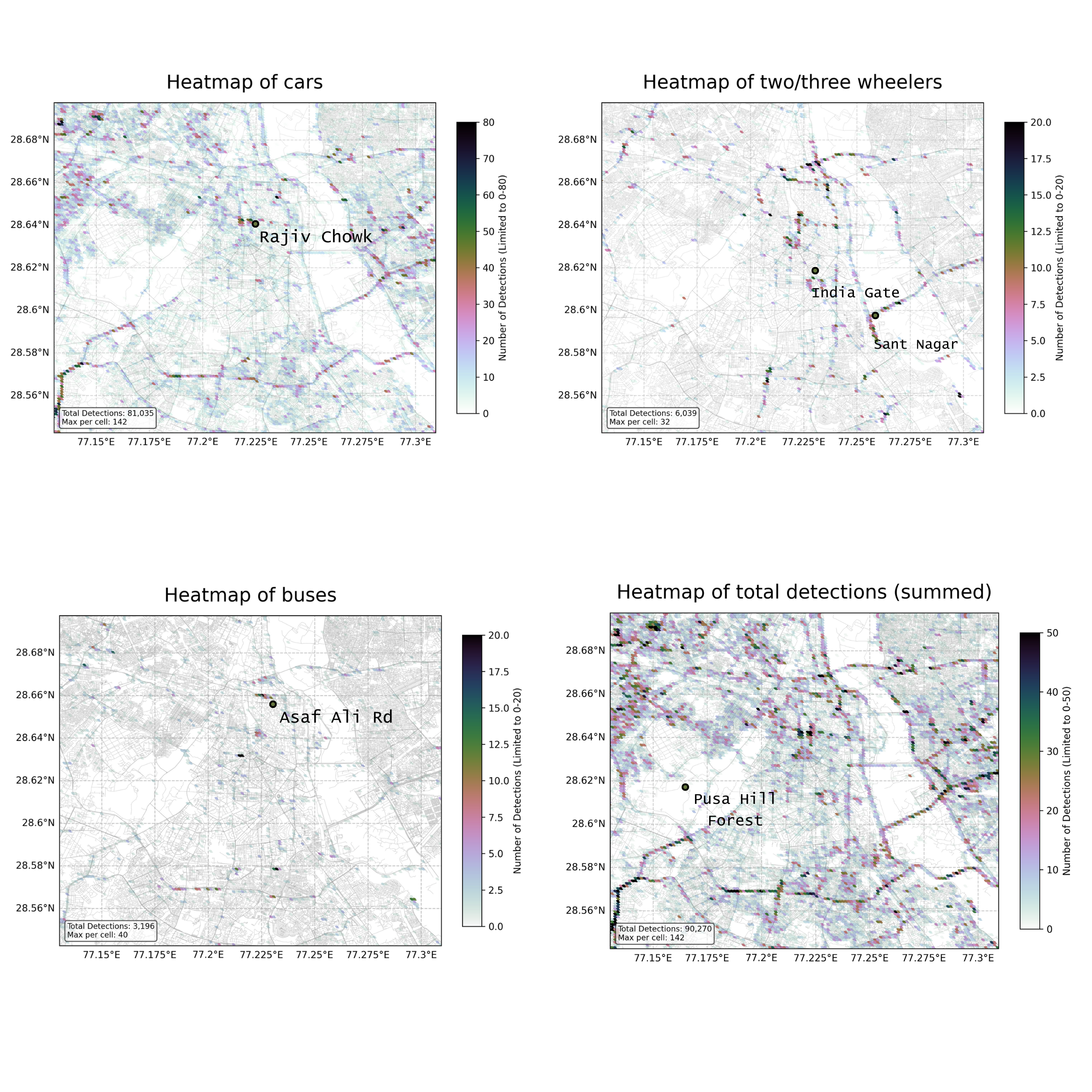}}
	\hspace{0.5cm}
        \caption{Heatmap of vehicles over A.O.I.}
\end{figure}

With a steady stream of live satellite data, the method could evolve into a robust real-time vehicular density algorithm. Technological advancements have rapidly improved models, with YOLOv8 evolving into YOLOv10 within just eight months. The latest developments indicate the forthcoming release of YOLOv11 and YOLOv12, which promise even better architecture, reduced latency, and overall improved performance, making the implementation even more promising for future applications.

\section{Conclusion}
This study underscores the potential of integrating AI with satellite imagery for real-time emission monitoring in urban settings. The approach is highly scalable and adaptable to various urban areas, offering valuable insights into traffic patterns, emissions, and air quality. The findings are particularly advantageous for urban planners, policymakers, and environmental agencies, providing a detailed tool for monitoring and mitigating vehicular emissions at a granular level.

The flexibility of this method comes from its ability to detect different object classes (e.g., industrial sources, brick kilns) and its adaptability to diverse regions. Additionally, it can incorporate more advanced machine learning models to improve accuracy. Future developments could involve integrating this inventory with air quality models to predict pollution levels based on real-time traffic data. Data collection for the study was facilitated using a QGIS plugin connected to Google satellite imagery. With continuous live satellite data, this approach has the potential to evolve into a robust real-time vehicular density algorithm. Rapid technological advancements have significantly enhanced models, with YOLOv8 advancing to YOLOv10 in just eight months. Looking ahead, the anticipated release of YOLOv11 and YOLOv12 promises further improvements in architecture, reduced latency, and overall performance, making this methodology even more promising for future applications.

\section*{Acknowledgements}
We would like to express our sincere gratitude to the Indian Institute of Tropical Meteorology and The University of Texas at Austin for their invaluable support in facilitating this project. This work was made possible through the dedicated collaboration of professionals and staff at both institutions, whose assistance ensured the smooth progression of our research. We also extend our appreciation to the open-source communities, online forums, and resources that provided essential support throughout the study. Finally, we are deeply thankful to our friends and families for their unwavering encouragement and understanding during the course of this work.

\section*{Ethical Statement}
The authors affirm that this research was conducted in an ethical and responsible manner. All satellite data and imagery used in the study were obtained from publicly available sources and used in compliance with applicable regulations and permissions. No human or animal subjects were involved in this research, thus there were no concerns regarding their welfare or privacy. Furthermore, the study adhered to all institutional and international ethical guidelines for data analysis and environmental research. The authors declare no conflict of interest related to this work, and no funding sources exerted undue influence on the outcomes or interpretations presented.

\pagebreak

\section*{References}
\begin{itemize}
    \item Poonam Mangaraj, Saroj Kumar Sahu, and Gufran Beig. Development of emission inventory for air quality assessment and mitigation strategies over most populous Indian megacity, Mumbai. \textit{Urban Climate}, 55:101928, 2024.
    \item Chandra Venkataraman, Arushi Sharma, Kushal Tibrewal, Suman Maity, and Kaushik Muduchuru. Carbonaceous aerosol emissions sources.
    \item Chinmay Jena, Sachin D Ghude, Rajesh Kumar, Sreyashi Debnath, Gaurav Govardhan, Vijay K Soni, Santosh H Kulkarni, G Beig, Ravi S Nanjundiah, and M Rajeevan. Performance of high resolution (400 m) PM2.5 forecast over Delhi. \textit{Scientific Reports}, 11(1):4104, 2021.
    \item Ao Wang, Hui Chen, Lihao Liu, Kai Chen, Zijia Lin, Jungong Han, and Guiguang Ding. YOLOv10: Real-time end-to-end object detection. arXiv preprint arXiv:2405.14458, 2024. Available at: \textcolor{blue}{\url{https://arxiv.org/abs/2405.14458}}
    \item Carly Sakumura. Seeing a better world from space. 2019.
    \item Zhu, Li, Xiong, Tianyu, Wang, Hongfeng, "YOLOv10: Real-Time End-to-End Object Detection", 2024. Available at: \textcolor{blue}{\url{https://arxiv.org/abs/2405.14458}}
    \item Redmon, Joseph, Divvala, Santosh, Girshick, Ross, Farhadi, Ali, "You Only Look Once: Unified, Real-Time Object Detection", 2015. Available at: \textcolor{blue}{\url{https://arxiv.org/abs/1506.02640}}
    \item Boukamcha, Hamdi, "YOLOv10 vs YOLOv8: A Comparative Analysis", 2023. Available at: \textcolor{blue}{\url{https://medium.com/@boukamchahamdi/yolov10-vs-yolov8-a-comparative-analysis-dc7b870a3331}}
    \item Ultralytics, "Ultralytics YOLO GitHub Repository", 2023. Available at: \textcolor{blue}{\url{https://github.com/ultralytics}}
\end{itemize}

\bibliography{references}

\end{document}